\documentclass[letterpaper, 10 pt, journal, twoside]{ieeetran}
\IEEEoverridecommandlockouts

\usepackage{graphicx}
\usepackage{amsmath, amsfonts} 
\usepackage{hyperref}
\usepackage{xcolor}
\usepackage{caption}
\usepackage{subcaption}

\title{Range-Visual-Inertial Odometry: Scale Observability Without Excitation}

\author{Jeff Delaune,
        David S. Bayard
        and Roland Brockers
\thanks{Manuscript received: October, 15, 2020; Revised January, 12, 2021; Accepted February, 5, 2021.}
\thanks{This paper was recommended for publication by Javier Civera upon evaluation of the Associate Editor and Reviewers' comments.
The research described in this paper was carried out at the Jet Propulsion Laboratory, California Institute of Technology, under a
contract with the National Aeronautics and Space Administration (80NM0018D0004). \textcopyright~2021 California Institute of Technology.
Government sponsorship acknowledged.}
\thanks{The authors are with the Jet Propulsion Laboratory,
        California Institute of Technology, Pasadena, CA.
        {\{jeff.h.delaune, david.s.bayard, roland.brockers\}@jpl.nasa.gov}}
\thanks{Digital Object Identifier (DOI): 10.1109/LRA.2021.3058918.}
}

\begin{document}

\thispagestyle{empty}

\onecolumn
{
\centering
This paper has been accepted for publication in \emph{IEEE Robotics and Automation Letters}, with presentation at the \emph{IEEE International Conference on Robotics and Automation (ICRA)}, Xi’an, China, 2021.\linebreak

DOI: \href{https://ieeexplore.ieee.org/document/9353193}{10.1109/LRA.2021.3058918}\linebreak
IEEE Xplore: \url{https://ieeexplore.ieee.org/document/9353193}\linebreak

© 2021 IEEE. Personal use is permitted, but republication/redistribution requires IEEE permission.
See \url{https://www.ieee.org/publications/rights/index.html} for more information.}

\twocolumn
\newpage
\maketitle
\setcounter{page}{1}

\begin{abstract}

Traveling at constant velocity is the most efficient trajectory for most robotics applications. Unfortunately
without accelerometer excitation, monocular Visual-Inertial Odometry (VIO) cannot observe scale and suffers severe error
drift. This was the main motivation for incorporating a 1D laser range finder in the navigation system for NASA’s
\emph{Ingenuity} Mars Helicopter. However, \emph{Ingenuity}'s simplified approach was limited to flat terrains.
The current paper introduces a novel range measurement update model based on using facet constraints. The
resulting range-VIO approach is no longer limited to flat scenes, but extends to any arbitrary structure for
generic robotic applications. An important theoretical result shows that scale is no longer in the right nullspace of
the observability matrix for zero or constant acceleration motion. In practical terms, this means that scale becomes
observable under constant-velocity motion, which enables simple and robust autonomous operations over arbitrary terrain.
Due to the small range finder footprint, range-VIO retains the minimal size, weight, and power attributes of VIO, with
similar runtime. The benefits are evaluated on real flight data representative of common
aerial robotics scenarios. Robustness is demonstrated using indoor stress data and full-state ground truth. We release our
software framework, called xVIO, as open source.

\end{abstract}

\begin{IEEEkeywords}
Visual-Inertial SLAM, Aerial Systems: Perception and Autonomy, Observability, Inertial Excitation, Mars Helicopter.
\end{IEEEkeywords}


\section{Introduction}

\IEEEPARstart{M}{onocular} Visual-Inertial Odometry (VIO) is a popular approach in robotics to obtain accurate
metric state estimates close to a scene, or in GPS-denied conditions. Indeed, a camera and an Inertial Measurement Unit (IMU)
form a minimal sensor suite in terms of size, weight and power, which is readily available on most robots.

However, monocular VIO can only observe the motion scale when the acceleration is not constant.
This leads to severe error drift under zero or constant-velocity trajectories, which are very common in robotics. This problem is critical
problem for applications which must rely on accurate VIO scale
estimates for control. Our work is motivated by Mars helicopters~\cite{bayard2019scitech,delaune2020}, but it is applicable to planetary,
military, and urban robots in general; as well as indoor or underground traverses along a straight corridor or tunnel. 

Our novel range-visual-inertial odometry algorithm can observe scale even under
zero or constant-acceleration trajectories. It uses a 1D Laser Range Finder (LRF) that keeps the sensor suite
lightweight, while efficiently leveraging VIO sparse structure estimates. Our main
contributions are:
\begin{itemize}
    \item a range measurement model which prevents VIO scale drift and adapts to any scene structure, 
    \item a linearized range-VIO observability analysis, showing scale is observable without excitation,
    \item outdoor demonstration on a realistic dataset,
    \item indoor stress case analysis with full-state ground truth, 
    \item an open-source C++ implementation.
\end{itemize}

In \cite{bayard2019scitech}, a range-VIO method was presented that navigates over relatively flat terrain while supporting a stable motionless hover needed for demonstrating NASA's \emph{Ingenuity} Mars Helicopter. The current paper extends these range-VIO results with a new method that makes scale observable for 3D terrain without
requiring any inertial excitation. This generalization addresses an important need in the field of robotics as well as for future Mars helicopters. The current paper is a journal extension of a previous 
conference paper~\cite{delaune2020}, which focused specifically on the Mars helicopter application. This included a real-time demonstration with candidate spaceflight hardware operating over Mars-like terrain. The conference paper treatment was non-theoretical and focused on obtaining proof-of-concept empirical results. The current journal paper derives and analyzes its theoretical observability properties. The error drift reduction is evaluated on urban aerial robotics data, which is significantly more complex and 3D than a Mars environment. The robustness of the facet-scene assumption is demonstrated with an indoor stress test supported by a full-state ground truth comparison. Finally, we make the source code publicly available\footnote{\href{https://github.com/jpl-x}{https://github.com/jpl-x}}.

We refer to monocular VIO as VIO in the rest of this paper.
Section~\ref{sec:litrev} reviews the VIO literature, observability limitations, and drift mitigation techniques using
additional sensors. Section~\ref{sec:rvio} introduces our range-VIO framework. Section~\ref{sec:observability} analyzes the observability benefits of our approach. Sections~\ref{sec:testing} and \ref{sec:results} present our tests and results. A video and a technical
report~\cite{delaune2020b} supplement this paper.

\section{Literature Review}
\label{sec:litrev}

\subsection{Visual-Inertial Odometry}

One branch of VIO is based on loosely-coupled visual-inertial sensing. In these approaches,
a vision-only algorithm estimates position and velocity up to scale, and orientation up to gravity, before fusion with the IMU~\cite{weiss2012icra}.
The visual odometry module can be swapped between any of the modern algorithms developed in the computer vision community, such
as PTAM~\cite{klein2007ismar}, SVO~\cite{forster2014icra}, ORB-SLAM~\cite{murTRO2015} or DSO~\cite{engel2018pami}.

The most accurate and robust VIO methods come from tightly-coupled approaches, in which visual measurements consisting
of feature tracks or image patch intensities directly constrain the inertial state integration in one single estimator.
These approaches require a larger state vector, which leads to a higher computational cost. But they gain in accuracy through the cross-correlations
between the inertial and visual states~\cite{leutenegger2014ijrr}, and in robustness with the ability to propagate the state even when no or few image primitives
can be tracked. Recent approaches include both filter-based~\cite{mourikis2007icra, Bloesch2015} and nonlinear
optimization-based methods~\cite{forster2017ieee, Qin2017, Stumberg2018}. Some solutions use image feature
coordinates for measurements~\cite{mourikis2007icra, forster2017ieee, Qin2017} while others use image intensity
values~\cite{Bloesch2015, Stumberg2018}. With good excitation, typical position errors can be under $1\%$ of distance travelled~\cite{delmerico2018icra}.

\subsection{VIO Observability Analysis}

VIO observability with unknown IMU bias has been studied at length in the robotics literature. Under generic excitation, the VIO states were found
to be observable except for the global position and the rotation about the gravity vector. \cite{martinelli2011tr, eagle2011ijrr, kelly2011ijrr} demonstrated this for the nonlinear system; while \cite{li2013ijrr, heschtro2014} proved it for the linearized system and improved its consistency. These unobservable quantities imply that VIO position and heading estimates drift under any conditions with noise. In practice, this drift is acceptable in many robotics scenarios operating at small scale.

\cite{wu2016minn} further analyzed the unobservable directions under two specific motions for the linearized system with unknown bias. First,
they showed that all three global rotations become unobservable if the system has no rotational motion of its own. Second, they showed that
under constant acceleration, the scale of motion is unobservable. \cite{kottas2013iros} derived results in line with these for the specific
case of hovering. The complete absence of rotation is unlikely in most real applications, and even if it happens, the relative orientation of the camera with respect
to the scene structure is still preserved. Constant or zero-acceleration is likely along straight traverses though, and the consequences
of scale errors in terms of position and velocity drift can be catastrophic for the planning and control of a robot's trajectory. 

\subsection{VIO Scale Drift Mitigation}

Range sensors and their equivalents can be used in addition to, or in replacement of, a monocular camera, in order to eliminate the scale observability issue of VIO. Most approaches
leverage either lidar or radar scans~\cite{mohamed2019ieee}, RGBD cameras~\cite{angladon2017jrtip}, or stereo visual measurements~\cite{taketomi2017cva}.
Unlike VIO, these options suffer either from range limitation or integration costs\footnote{E.g. lidar scanner weight, or stereo baseline.} that limit their use in robotics applications.

1D Laser Range Finders (LRF) are an underrepresented sensing option in the SLAM literature. Modern units can sense over tens of meters with centimeter resolution. They fit within a small, lightweight and power-efficient package that can be accommodated even on resource-constrained robots. In our previous work for NASA's \emph{Ingenuity} Mars Helicopter~\cite{bayard2019scitech}, we implemented a range-visual-inertial odometry algorithm that integrates LRF measurements to make the scale observable. The low dimension of the resulting estimator, only 21 states, comes at the cost of assuming the scene is flat and level, which is not compatible with most robotics scenarios. \cite{urzua2017mav} solve a similar problem over 3D scenes by initializing the depth of some VIO features with ultrasonic range measurements. This relaxes the scene assumption from globally-flat to locally-flat, but it also assumes the local terrain slope is perpendicular to the range sensor axis within the ranged area. This is problematic over 3D scenes,
given the large beam width of ultra sonic sensors.

In this paper, we eliminate VIO scale drift over any scene structure using a novel LRF measurement model. The accuracy and narrow beam width of the
LRF create a strong range constraint with the depth of the visual features estimated by VIO in
an Extended Kalman Filter (EKF). This constraint assumes the scene can be partitioned into triangular facets with the visual image features as vertices.

\section{Range-Visual-Inertial Odometry}
\label{sec:rvio}

\begin{figure}
\centering
\includegraphics[width=\columnwidth]{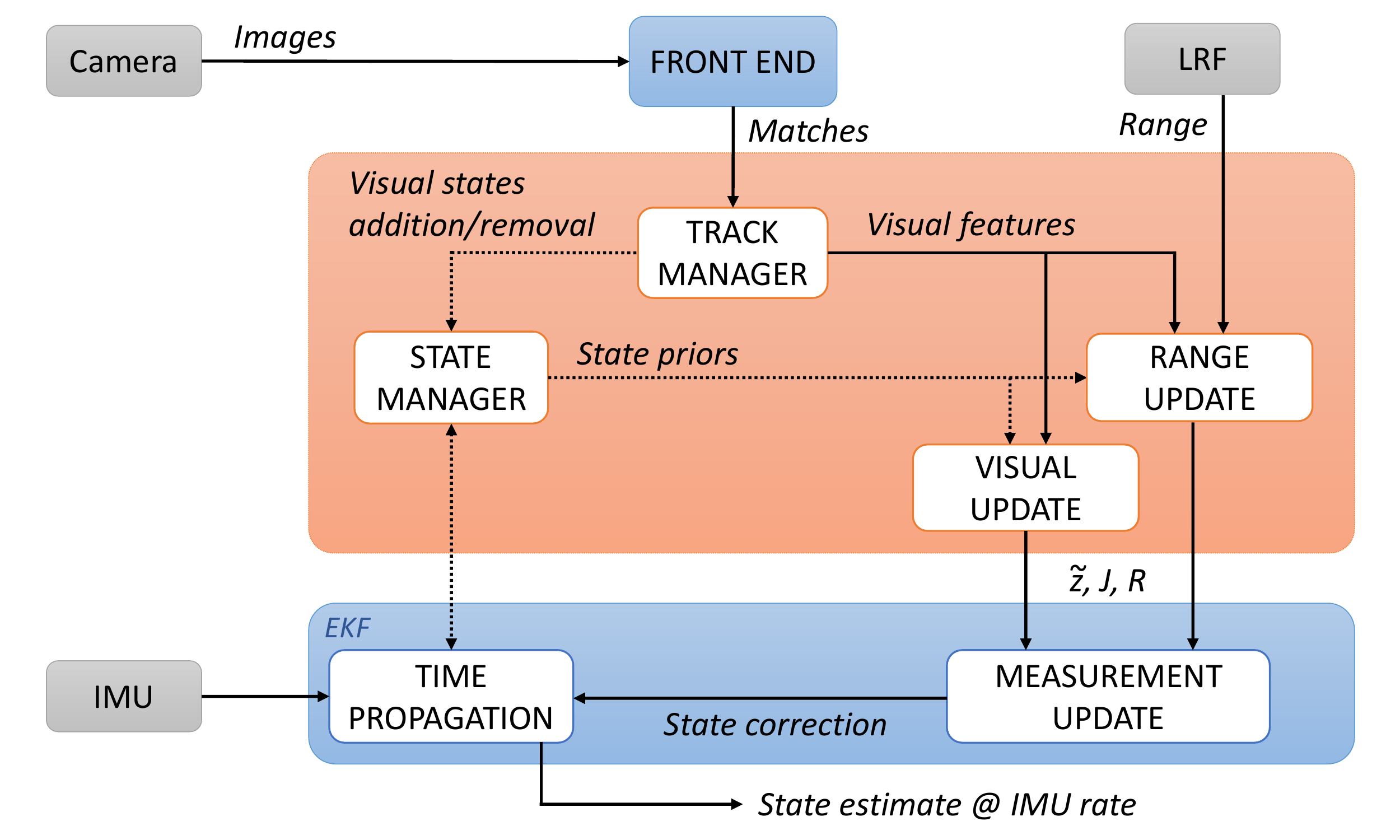}
\caption{Range-visual-inertial odometry architecture. Range and visual
measurement innovations $\tilde{\boldsymbol{z}}$, Jacobians $\boldsymbol{J}$ and
covariance matrices $\boldsymbol{R}$ are used to correct the inertial navigation errors in an EKF.
The track manager sorts image features matches into tracks, while the state manager adds and removes or vision states dynamically.}
\label{fig:rvio-architecture}
\end{figure}

The architecture of our framework in Figure~\ref{fig:rvio-architecture} is based on an Extended Kalman Filter (EKF).
It tightly couples visual and range updates with inertial state propagation. We provide complete derivation details in our technical report~\cite{delaune2020b}.

\subsection{Inertial State Propagation}

The EKF state vector $\boldsymbol{x}=\begin{bmatrix}{\boldsymbol{x}_{I}}^T &
{\boldsymbol{x}_{V}}^{T}\end{bmatrix}^{T}$ is divided between the states related to the IMU $\boldsymbol{x}_{I}$,
and those related to vision $\boldsymbol{x}_{V}$. The inertial states
\begin{equation}
\boldsymbol{x}_{I} =
\begin{bmatrix}
{\boldsymbol{p}_w^i}^T &
{\boldsymbol{v}_w^i}^T &
{\boldsymbol{q}_{w}^{i}}^T &
{\boldsymbol{b}_{g}}^{T} &
{\boldsymbol{b}_{a}}^{T}
\end{bmatrix}^{T}
\label{eq:state-vec-i}
\end{equation}
include the position, velocity and orientation of the IMU frame $\left\{\mathit{i}\right\}$ with respect to the world frame $\left\{\mathit{w}\right\}$, the gyroscope biases
$\boldsymbol{b}_{g}$ and the accelerometer biases $\boldsymbol{b}_{a}$. Rotation quaternions are used to model orientations.

IMU measurements are used to propagate the state estimate and the corresponding subblocks of the error covariance
matrix to first order, using \cite{weiss2011icra} and \cite{trawny2005minn}.

\subsection{Visual Update}

We perform visual updates using the hybrid SLAM-MSCKF paradigm~\cite{li2012rss}. This requires additional vision states
\begin{equation}
\begin{aligned}
&
\boldsymbol{x}_{V} = \left[\begin{matrix}
{\boldsymbol{p}_w^{c_1}}^T & ... & {\boldsymbol{p}_w^{c_M}}^T &
{\boldsymbol{q}_{w}^{c_1}}^T & ... & {\boldsymbol{q}_{w}^{c_M}}^T \end{matrix}\right.
\\
& \qquad\qquad\qquad\qquad\qquad\qquad\qquad\quad
\left.\begin{matrix}
{\boldsymbol{f}_{1}}^{T} & ... & {\boldsymbol{f}_N}^T
\end{matrix}\right]^T
\end{aligned}
\label{eq:state-vec-v}
\end{equation}
which includes a sliding window with the orientations $\left\{\boldsymbol{q}_{w}^{c_i}\right\}_{i}$
and positions $\left\{\boldsymbol{p}_w^{c_i}\right\}_{i}$ of the camera frame at the last $M$ image
time instances, along with the 3D coordinates of $N$ visual features $\left\{\boldsymbol{f}_{j}\right\}_{j}$.
Each feature state
$\boldsymbol{f}_j = \begin{bmatrix}
\alpha_j &
\beta_j &
\rho_j
\end{bmatrix}^{T}$
represents the inverse-depth parametrization of world feature
point $\boldsymbol{F}_j$ with respect to an anchor pose $\left\{c_{i_j} \right\}$
selected from the sliding window of pose states. Inverse depth improves feature
depth convergence properties~\cite{civera2008tro}.

The visual measurement is the pinhole projection of terrain feature $\boldsymbol{F}_j$ over
the normalized image plane $z=1$ of the camera frame $\left\{c_i\right\}$ at time $i$
\begin{equation}
^{i,j}\boldsymbol{z}_{v,m} =
\frac{1}{^{c_i}z_j}
\begin{bmatrix}^{c_i}x_j\\^{c_i}y_j\end{bmatrix}
+ {\boldsymbol{n}_v}\;,
\label{eq:vision-measurement}
\end{equation}
where $\boldsymbol{n}_v$ is a zero-mean white Gaussian feature measurement noise in
image space. Equation~\ref{eq:vision-measurement} can be related to the state if we
express the Cartesian coordinates of feature $\boldsymbol{F}_j$ in camera frame
$\left\{c_i\right\}$ as
\begin{align}
{\boldsymbol{p}_{c_i}^{F_j}} &= \begin{bmatrix}
^{c_i}x_j & ^{c_i}y_j & ^{c_i}z_j \end{bmatrix}^T
\label{eq:feature-cart-coord}\\
&= \boldsymbol{C}(\boldsymbol{q}_w^{c_i})
\Bigg(\boldsymbol{p}_w^{c_{i_j}} +
\frac{1}{\rho_j} \boldsymbol{C}(\boldsymbol{q}_w^{c_{i_j}})^T
\begin{bmatrix} \alpha_j \\ \beta_j \\ 1\end{bmatrix}
- \boldsymbol{p}_w^{c_i}\Bigg)
\label{eq:feature-frame-tf1}
\end{align}
This enables SLAM updates for features which are included in the state vector~\cite{delaune2019iros}.
Features which are not included in the state vector are processed using MSCKF~\cite{mourikis2007icra}.
MSCKF updates have a linear cost per feature, as opposed to a cubic cost for SLAM. However MSCKF requires
translational motion since the feature has to be triangulated, which is not always satisfied in practice. Hence, we always
perform SLAM updates, and only use MSCKF when the translational motion allows for it. This hybrid approach is also the most computationally-efficient~\cite{li2012rss}. SLAM features are either initialized with semi-infinite
depth uncertainty~\cite{civera2008tro}, or using a MSCKF prior if possible~\cite{li2012rss}.

Visual corner features are detected in the image using the FAST algorithm~\cite{rosten2010faster}, and tracked with
the pyramidal implementation of the Kanade-Lucas-Tomasi algorithm~\cite{Bouguet2000, shi1994}. Outlier features
are detected at two levels: first at the image level with RANSAC~\cite{fischler1981acm}, and then at the filter level with
a Mahalanobis distance test. The track manager module in Figure~\ref{fig:rvio-architecture} assigns each feature
to either the SLAM or MSCKF paradigm based on the track length, detection score, and image coordinates. We use
image tiles to ensure SLAM features are distributed throughout the field of view, and ensure strong pose constraints.

\subsection{Ranged Facet Update}

Our main contribution is a novel range measurement model to constrain the VIO scale drift. Like VIO, it is designed to work over arbitrary unknown 2D or 3D scenes.

\subsubsection{Measurement Model}

A range measurement depends on both the pose of the range sensor and the structure of the scene. The associated measurement
model in a Bayesian estimator should account for uncertainty on both. Since structure uncertainty is included in the
SLAM feature states, we leverage these states to construct the new range update model.

\begin{figure}[h]
\centering
\includegraphics[width=2.4in]{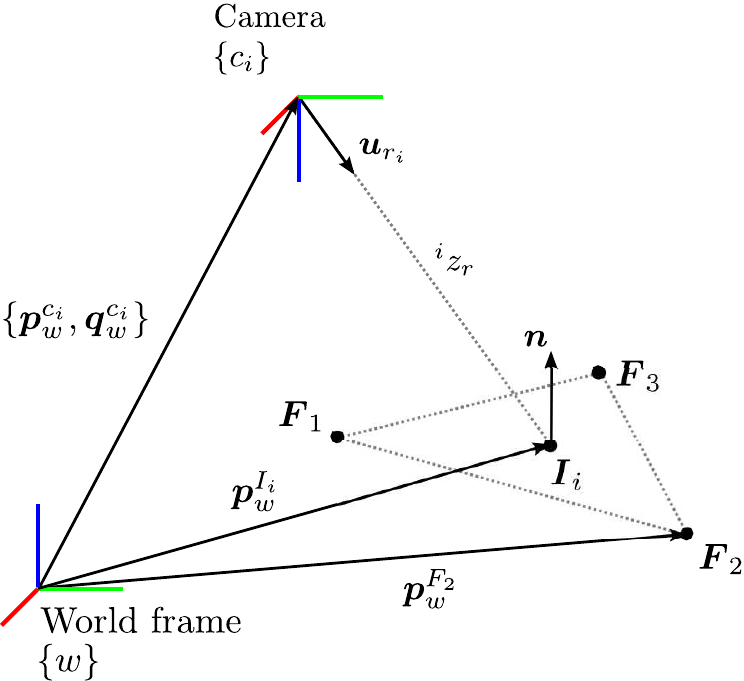}\\
\caption{Geometry of the range measurement ${^i{z}}_r$ at time $i$. The scene is assumed to be locally flat within a facet formed by
visual features $\boldsymbol{F}_1$, $\boldsymbol{F}_2$ and $\boldsymbol{F}_3$ to build the range constraint.}
\label{fig:rvio-geometry}
\end{figure}

Our key assumption is that the structure is locally flat between three SLAM features surrounding the intersection point of the LRF
beam with the scene. This assumption derives from the observation that visual features are often located at depth discontinuities, and
that the structure of the scene between features is often smooth. The impact of this assumption in real-world sequences is discussed in the results section. For simplification
purposes in this paper, we also assume zero translation between the optical center of the camera and the origin of the LRF\footnote{This
offset can be measured and introduced in the model if needed.}.
Figure~\ref{fig:rvio-geometry} illustrates the geometry of the scene.
$\boldsymbol{u}_{r_i}$ is the unit vector oriented along the optical axis of the
LRF at time $i$. $\boldsymbol{I_i}$ is the intersection of this axis with the terrain.
$\boldsymbol{F}_1$, $\boldsymbol{F}_2$ and $\boldsymbol{F}_3$ are
SLAM features forming a triangle around $\boldsymbol{I_i}$ in image space.
$\boldsymbol{n}$ is a normal vector to the plane containing
$\boldsymbol{F}_1$, $\boldsymbol{F}_2$, $\boldsymbol{F}_3$ and
$\boldsymbol{I_i}$.

If the dot product ${{\boldsymbol{u}_{r_i}}}\cdot{\boldsymbol{n}} \neq 0$, we can express the range measurement at time $i$ as
\begin{align}
^i{z}_r &=
{^i{z}_r} \frac{\boldsymbol{u}_{r_i}\cdot\boldsymbol{n}}
	{\boldsymbol{u}_{r_i}\cdot\boldsymbol{n}}
\label{eq:rvio-1}
\\
&=
\frac{(\boldsymbol{p}_w^{I_i} - \boldsymbol{p}_w^{c_i})\cdot\boldsymbol{n}}
	{\boldsymbol{u}_{r_i}\cdot\boldsymbol{n}}
\label{eq:rvio-2}\\
&=
\frac{(\boldsymbol{p}_w^{I_i} - \boldsymbol{p}_w^{F_2} + \boldsymbol{p}_w^{F_2} - \boldsymbol{p}_w^{c_i})\cdot\boldsymbol{n}}
		{\boldsymbol{u}_{r_i}\cdot\boldsymbol{n}}
\\
&=
\frac{(\boldsymbol{p}_w^{F_2} - \boldsymbol{p}_w^{c_i})\cdot\boldsymbol{n}}
		{\boldsymbol{u}_{r_i}\cdot\boldsymbol{n}}
\label{eq:rvio-3}
\end{align}
since $(\boldsymbol{p}_w^{I_i} - \boldsymbol{p}_w^{F_2})\cdot\boldsymbol{n} = 0$, where
\begin{equation}
\boldsymbol{n}= (\boldsymbol{p}_w^{F_1} - \boldsymbol{p}_w^{F_2})
	\times(\boldsymbol{p}_w^{F_3} - \boldsymbol{p}_w^{F_2})
\label{eq:facet-normal}
\end{equation}
Here, $\left\{\mathit{c_i}\right\}$ is the camera frame at time $i$, and
$\boldsymbol{p}_w^{*}$ represents the position of an object in the world frame
$\left\{\mathit{w}\right\}$. Note that $\boldsymbol{n}$ is not necessarily a unit
vector in this analysis.

Equations~(\ref{eq:rvio-3}) and (\ref{eq:facet-normal}) demonstrate that range can be expressed as a nonlinear
function $ h_r$ of the state vector $\boldsymbol{x}$ in Equation~(\ref{eq:rvio-meas}), without requiring any additional state beyond those of VIO. For use in our EKF estimation framework, we assume that the LRF measurements are disturbed by additive zero-mean white Gaussian noise $n_r$.
\begin{align}
{^i}z_{r,m}
&= {^iz_r} + {n_r}
= h_r(\boldsymbol{x}) + {n_r}
\label{eq:rvio-meas}
\end{align}
We refer the reader to ~\cite{delaune2020b} for the linearization of Equation~\ref{eq:rvio-meas}, providing the expressions of the measurement Jacobians.

\subsubsection{Delaunay triangulation}

To construct the range update in practice, we perform a Delaunay triangulation in image space
over the SLAM features, and select the triangle in which the intersection of the LRF beam
with the scene is located. We opted for the Delaunay triangulation since it maximizes the smallest angle of all possible
triangulations~\cite{gartner2013ETH}. This property avoids ``long and skinny" triangles that
do not provide strong local planar constraints.

\begin{figure}[h]
\centering
\includegraphics[width=0.92\columnwidth]{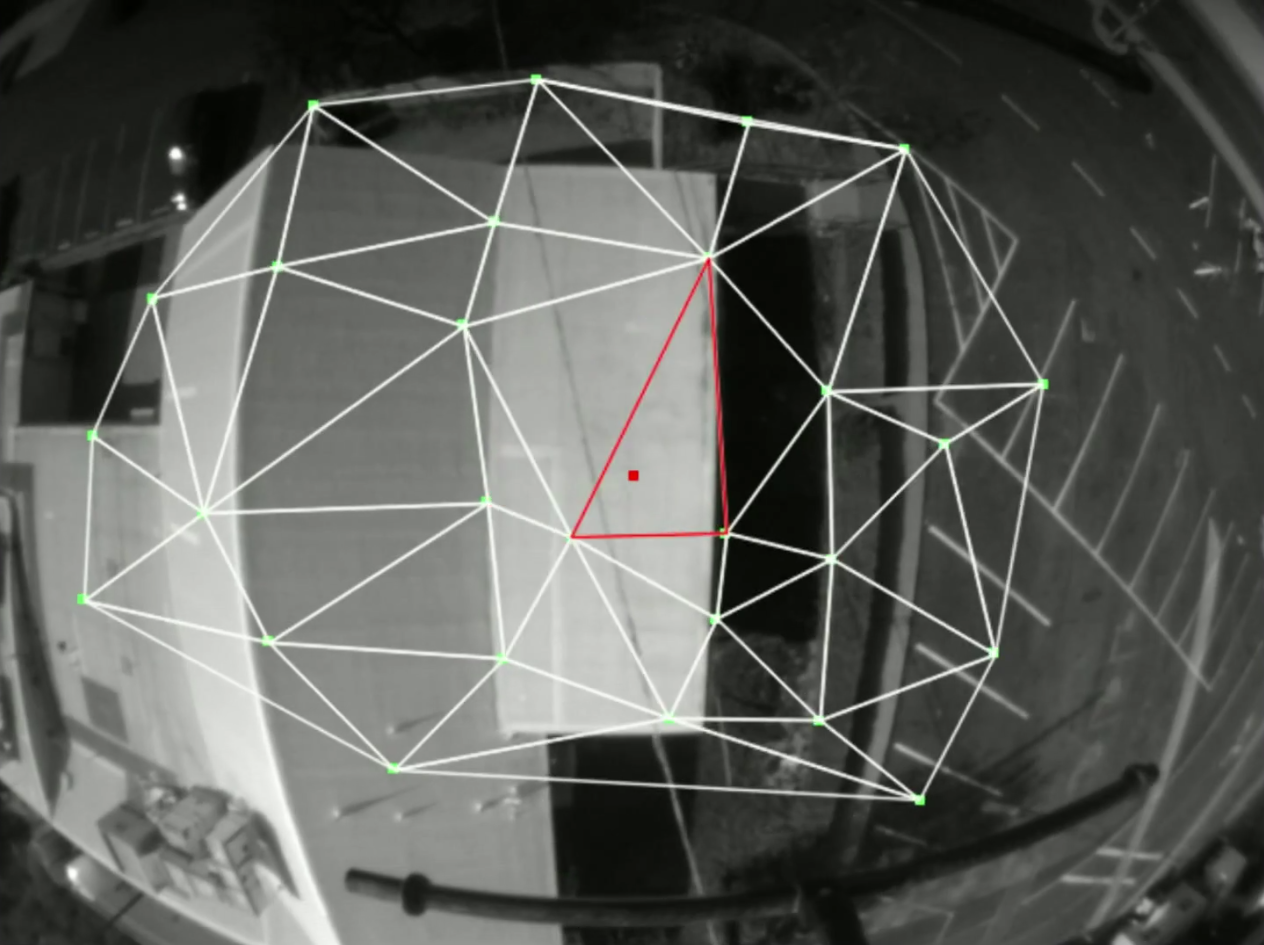}\\
\caption{Delaunay triangulation between SLAM image features tracked in the outdoor flight dataset. The red dot
represents the intersection point of the LRF beam and the surface. The surrounding red triangle
is the ranged facet.}
\label{fig:delaunay}
\end{figure}

Figure~\ref{fig:delaunay} shows the Delaunay triangulation, and
the triangle selected as a ranged facet, over a sample image from our outdoor test sequence.
It also illustrates the partitioning of the scene into triangular facets, with SLAM features at their corners. Note
that if the state estimator uses only 3 SLAM features in a lightweight fashion,
this is equivalent to a globally-flat world assumption. Conversely, if the density of SLAM features
increases, the areas of the facets tend to zero and the facet scene assumption virtually
disappears.

\subsubsection{Range outlier rejection}

Before being used in the filter, the range measurements go through a Mahalanobis distance test to detect outliers. This gating compares
the range measurement to a prior built from the coordinates of the three visual features in the facet. It rejects violations
of the facet assumption that cannot be explained by the prior uncertainty model derived from the error covariance matrix.

\section{Observability Analysis}
\label{sec:observability}

We perform the observability analysis of the linearized range-VIO system, since it is based on an EKF. Although the observability of
the nonlinear system would be required for completeness, it is out of the scope of this paper.

To simplify the equations, our analysis assumes a state vector $\boldsymbol{x}^0=\begin{bmatrix}{\boldsymbol{x}_{I}}^T &
{\boldsymbol{x}_P}^{T}\end{bmatrix}^{T}$, where $\boldsymbol{x}_{I}$ was defined in Equation~(\ref{eq:state-vec-i})
and
$\boldsymbol{x}_P =
{\begin{bmatrix}
{^w\boldsymbol{p}_1}^T & ... & {^w\boldsymbol{p}_N}^T
\end{bmatrix}}^T$
includes the Cartesian coordinates of the $N$ SLAM features, $N \geq 3$. \cite{li2013ijrr} proved that observability analysis for the linearized system
based on $\boldsymbol{x}^0$ is equivalent to observability analysis for the linearized system defined with $\boldsymbol{x}$ in the previous section.

\subsection{Observability Matrix}

For $k\geq1$, $\boldsymbol{M}_k = \boldsymbol{H}_k \boldsymbol{\Phi}_{k,1}$ is the k-th block row of observability matrix
$\boldsymbol{M}$. $\boldsymbol{H}_k$ is the Jacobian of the range measurement in Equation~(\ref{eq:rvio-3}) at time $k$
with respect to $\boldsymbol{x}^0$, which is derived in Equations~(42-47) in~\cite{delaune2020b}. $\boldsymbol{\Phi}_{k,1}$
is the state transition matrix from time $1$ to time $k$~\cite{hesch2012minn}.

Without loss of generality, we can assume the ranged facet is constructed from the first 3 features in $\boldsymbol{x}_P$\footnote{
This ordering of the states can be obtained at any timestep using permutation matrices. Permutation matrices are full rank and
hence do not affect the rank of the observability matrix.}. Then we derive
the following expression for $\boldsymbol{M}_k$ in~\cite{delaune2020b}.
\begin{equation}
    \begin{aligned}
        &
        \boldsymbol{M}_k = \frac{1}{b}\big[
        \begin{matrix}
            \begin{array}{c|c|c|c|c|c|c}
                \boldsymbol{M}_{k,p} &
                \boldsymbol{M}_{k,v} &
                \boldsymbol{M}_{k,q} &
                \boldsymbol{M}_{k,b_g} &
                \boldsymbol{M}_{k,b_a}
            \end{array}
        \end{matrix}\\
        &\quad\quad\quad\quad\quad
        \begin{matrix}
            \begin{array}{c|c|c|c}
                \boldsymbol{M}_{k,p_1} &
                \boldsymbol{M}_{k,p_2} &
                \boldsymbol{M}_{k,p_3} &
                \boldsymbol{0}_{1\times3(N-3)}
            \end{array}
        \end{matrix}\big]
    \end{aligned}
\label{eq:obs-matrix}
\end{equation}
where
\begin{align}
    \boldsymbol{M}_{k,p} &= - {^w}\boldsymbol{n}^T
    \label{eq:obs-matrix-p}
    \\
    \boldsymbol{M}_{k,v} &= - (k-1) \delta t {^w}\boldsymbol{n}^T
    \label{eq:obs-matrix-v}
    \\
    &\mkern-50mu\begin{aligned}
        \boldsymbol{M}_{k,\theta} &= {^w}\boldsymbol{n} \Big( - \frac{a}{b} \boldsymbol{C}\left(\boldsymbol{q}_w^{c_k}\right)^T \left\lfloor {^c\boldsymbol{u}_r} \times \right\rfloor \boldsymbol{C}\left(\boldsymbol{q}_w^{i_k}\right)
        \\
        & - \Bigl\lfloor \boldsymbol{p}_w^{i_1} - \boldsymbol{v}_w^{i_1} (k-1) \delta t - \frac{1}{2}{^w}\boldsymbol{g}(k-1)^2 \delta t^2
        \\
        & \quad - \boldsymbol{p}_w^{i_k}\times \Bigr\rfloor \Big) \boldsymbol{C}\left(\boldsymbol{q}_{i_1}^w\right)
    \end{aligned}
    \label{eq:obs-matrix-theta}
    \\
    \boldsymbol{M}_{k,b_g} &= - \frac{a}{b}  {^w}\boldsymbol{n}^T \boldsymbol{C}\left(\boldsymbol{q}_w^{c_k}\right)^T \lfloor {^c\boldsymbol{u}_r} \times\rfloor \boldsymbol{\phi}_{12}  - {^w}\boldsymbol{n} \boldsymbol{\phi}_{52}
    \label{eq:obs-matrix-bg}
\end{align}
\begin{align}
    \boldsymbol{M}_{k,b_a} &= - {^w}\boldsymbol{n}^T \boldsymbol{\phi}_{54}
    \label{eq:obs-matrix-ba}
    \\
    \boldsymbol{M}_{k,p_1} &= \left( \left\lfloor(\boldsymbol{p}_w^{F_{3}} - \boldsymbol{p}_w^{F_{2}})\times\right\rfloor
\left(\boldsymbol{p}_w^{F_{2}} - \boldsymbol{p}_w^{I_k} \right) \right)^T
    \label{eq:obs-matrix-f1}
    \\
    \boldsymbol{M}_{k,p_2} &= \left( {^w}\boldsymbol{n} + 
    \left\lfloor(\boldsymbol{p}_w^{F_{1}} - \boldsymbol{p}_w^{F_{3}})\times\right\rfloor
    \left(\boldsymbol{p}_w^{F_{2}} -\boldsymbol{p}_w^{I_k} \right) \right)^T 
    \label{eq:obs-matrix-f2}
    \\
    \boldsymbol{M}_{k,p_3} &= \left( \left\lfloor(\boldsymbol{p}_w^{F_{2}} - \boldsymbol{p}_w^{F_{1}})\times\right\rfloor \left(\boldsymbol{p}_w^{F_{2}} - \boldsymbol{p}_w^{I_k} \right) \right)^T
    \label{eq:obs-matrix-f3}
    \\
    a &= (\boldsymbol{p}_w^{f_{j_2}} - \boldsymbol{p}_w^{c_i})^T{^w\boldsymbol{n}}
    \label{eq:rv-aterm}
    \\
    b &= {{^w\boldsymbol{u}_{r_i}}^T}{^w\boldsymbol{n}}
\label{eq:rv-bterm}
\end{align}

and $\boldsymbol{\phi}_*$ are integral terms defined in \cite{hesch2012minn}.
For a generic vector $\boldsymbol{u} \in \mathbb{R}^3$, let
$^w\boldsymbol{u}$ represent its coordinates in the world frame $\left\{\mathit{w}\right\}$, and
$
\lfloor\boldsymbol{u}\times\rfloor = \begin{bmatrix}
0 & -u_{z} & u_{y}\\
u_{z} & 0 & -u_{x}\\
-u_{y} & u_{x} & 0\end{bmatrix}
$
the skew-symmetric matrix associated with it. $c_{i_1}$, $c_{i_2}$ and $c_{i_3}$ are the anchor poses
associated to $\boldsymbol{F}_1$, $\boldsymbol{F}_2$ and
$\boldsymbol{F}_3$, respectively, for their inverse-depth coordinates at time $i$.

\subsection{Unobservable Directions}

\subsubsection{Generic motion}

One can verify that the vectors spanning a global position or a rotation about the gravity vector still belong
to the right nullspace of $\boldsymbol{M}_k$. Thus, the ranged facet update does not improve the
observabily over VIO under generic motion~\cite{li2013
Code
Related Papers
About arXivLabs
ijrr}, which is intuitive. Likewise, in the absence of
rotation, the global orientation is still not observable~\cite{wu2016minn}.

\subsubsection{Constant acceleration}

In this subsection, we demonstrate that in the constant acceleration case, unlike VIO~\cite{wu2016minn}, the vector spanning
the scale dimension
\begin{equation}
\boldsymbol{N}_s =
\begin{bmatrix}
{\boldsymbol{p}_w^{i_1}}^T &
{\boldsymbol{v}_w^{i_1}}^T &
{\boldsymbol{0}_{6\times1}}^T &
- {^{i}\boldsymbol{a}_w^{i}}^T &
{\boldsymbol{p}_w^{F_1}}^T &
... &
{\boldsymbol{p}_w^{F_N}}^T
\end{bmatrix}^{T}
\label{eq:scale-null-vector}
\end{equation}
does not belong the right nullspace of $\boldsymbol{M}_k$, i.e. $\boldsymbol{M}_k \boldsymbol{N}_s \neq \boldsymbol{0}$.
${^{i}\boldsymbol{a}_w^{i}}$
is the zero or constant acceleration of the IMU frame in world frame, resolved in the IMU frame. One can write
\begin{align}
    &\begin{aligned}
    \boldsymbol{M}_k \boldsymbol{N}_s &=
        - {^w}\boldsymbol{n}^T {\boldsymbol{p}_w^{i_1}}
        - (k-1) \delta t {^w}\boldsymbol{n}^T {\boldsymbol{v}_w^{i_1}}
        + {^w}\boldsymbol{n}^T \boldsymbol{\phi}_{54} {^{i}\boldsymbol{a}_w^{i}}
        \\&
        + \left( \left\lfloor(\boldsymbol{p}_w^{F_{3}} - \boldsymbol{p}_w^{F_{2}})\times\right\rfloor \left(\boldsymbol{p}_w^{F_{2}} - \boldsymbol{p}_w^{I_k} \right) \right)^T {\boldsymbol{p}_w^{F_1}}
        \\&
        + \left( {^w}\boldsymbol{n} + \left\lfloor(\boldsymbol{p}_w^{F_{1}} - \boldsymbol{p}_w^{F_{3}})\times\right\rfloor \left(\boldsymbol{p}_w^{F_{2}} -\boldsymbol{p}_w^{I_k} \right) \right)^T {\boldsymbol{p}_w^{F_2}}
        \\&
        + \left( \left\lfloor(\boldsymbol{p}_w^{F_{2}} - \boldsymbol{p}_w^{F_{1}})\times\right\rfloor \left(\boldsymbol{p}_w^{F_{2}} - \boldsymbol{p}_w^{I_k} \right) \right)^T {\boldsymbol{p}_w^{F_3}}
    \end{aligned}
    \label{eq:mk-ns-1}
\end{align}
Code
Related Papers
About arXivLabs

Reference \cite{wu2016minn} shows that, under constant acceleration,
\begin{equation}
\boldsymbol{\phi}_{54} {^{i}\boldsymbol{a}_w^{i}} = - \left( {\boldsymbol{p}_w^{i_k}} - {\boldsymbol{p}_w^{i_1}} - (k-1)\delta t {\boldsymbol{v}_w^{i_1}} \right)
\end{equation}

so
\begin{align}
    &\mkern-60mu\begin{aligned}
    \boldsymbol{M}_k \boldsymbol{N}_s &=
        {^w}\boldsymbol{n}^T \left( \boldsymbol{p}_w^{F_2} - {\boldsymbol{p}_w^{i_k}}\right)
        \\&
        + \left(\boldsymbol{p}_w^{F_2} - \boldsymbol{p}_w^{I_k} \right)^T \Big( \left\lfloor(\boldsymbol{p}_w^{F_3} - \boldsymbol{p}_w^{F_2})\times\right\rfloor^T {\boldsymbol{p}_w^{F_1}}
        \\&
        + \left\lfloor(\boldsymbol{p}_w^{F_1} - \boldsymbol{p}_w^{F_3})\times\right\rfloor^T {\boldsymbol{p}_w^{F_2}}
        \\&
        + \left\lfloor(\boldsymbol{p}_w^{F_2} - \boldsymbol{p}_w^{F_1})\times\right\rfloor^T {\boldsymbol{p}_w^{F_3}}\Big)
    \end{aligned}
\end{align}
The cross product of the first term can be modified such that
\begin{align}
    &\begin{aligned}
        &\left(\boldsymbol{p}_w^{F_2} - \boldsymbol{p}_w^{I_k} \right)^T \left\lfloor(\boldsymbol{p}_w^{F_3} - \boldsymbol{p}_w^{F_2})\times\right\rfloor^T {\boldsymbol{p}_w^{F_1}} 
        \\
        &= \left(\boldsymbol{p}_w^{F_2} - \boldsymbol{p}_w^{I_k} \right)^T \left\lfloor(\boldsymbol{p}_w^{F_3} - \boldsymbol{p}_w^{F_2})\times\right\rfloor^T \left( {\boldsymbol{p}_w^{F_1} - \boldsymbol{p}_w^{I_k} + \boldsymbol{p}_w^{I_k}} \right)
    \end{aligned}
    \\
    &= \left(\boldsymbol{p}_w^{F_2} - \boldsymbol{p}_w^{I_k} \right)^T \left\lfloor(\boldsymbol{p}_w^{F_3} - \boldsymbol{p}_w^{F_2})\times\right\rfloor^T \boldsymbol{p}_w^{I_k}
\end{align}
By definition of the cross product,
\begin{equation}
\exists\,\lambda \in \mathbb{R}, \left\lfloor(\boldsymbol{p}_w^{F_3} - \boldsymbol{p}_w^{F_2})\times\right\rfloor^T \left( \boldsymbol{p}_w^{F_1}-\boldsymbol{p}_w^{I_k} \right) = \lambda {^w}\boldsymbol{n}
\end{equation}
and 
\begin{equation}
\lambda \left(\boldsymbol{p}_w^{F_2} - \boldsymbol{p}_w^{I_k} \right)^T {^w}\boldsymbol{n} = 0
\end{equation}
Thus, by applying this to all cross-product terms,
\begin{align}
    &
    \begin{aligned}
    		\boldsymbol{M}_k \boldsymbol{N}_s &= {^w}\boldsymbol{n}^T \left( \boldsymbol{p}_w^{F_2} - {\boldsymbol{p}_w^{i_k}}\right)
    		\\&
    		+ \left( \boldsymbol{p}_w^{F_2} - \boldsymbol{p}_w^{I_k} \right)^T \bigl\lfloor( \boldsymbol{p}_w^{F_3} - \boldsymbol{p}_w^{F_2} + \boldsymbol{p}_w^{F_1}
    		\\&
    		- \boldsymbol{p}_w^{F_3} + \boldsymbol{p}_w^{F_2} - \boldsymbol{p}_w^{F_1})\times\bigr\rfloor^T \boldsymbol{p}_w^{I_k}
    \end{aligned}
    \\
    &\mkern58mu
    = {^w}\boldsymbol{n}^T \left( \boldsymbol{p}_w^{F_2} - {\boldsymbol{p}_w^{i_k}}\right)
\end{align}
By definition, if the three features of the facet are not aligned in the image, ${^w}\boldsymbol{n}^T \left( \boldsymbol{p}_w^{F_2} - {\boldsymbol{p}_w^{i_k}}\right) \neq \boldsymbol{0}$. End of proof.

Unlike VIO,
range-VIO thus enables scale convergence even in the absence of acceleration excitation.

\subsubsection{Zero-velocity}

Note that when the velocity is null, i.e. in hover, ${\boldsymbol{v}_w^{i_1}} = \boldsymbol{0}$,
${\boldsymbol{p}_w^{i_1}} = {\boldsymbol{p}_w^{i_k}}$ in the previous demonstration, and the following unobservable direction appears instead.
\begin{equation}
\boldsymbol{N}_h =
\begin{bmatrix}
{\boldsymbol{0}_{24\times1}}^T &
{\boldsymbol{p}_w^{F_4}}^T &
... &
{\boldsymbol{p}_w^{F_N}}^T
\end{bmatrix}^{T}
\label{eq:partial-scale-hover-null-vector}
\end{equation}
It corresponds to the depth of the SLAM features not included in the facet. This result means that in
the absence of translation motion, when feature depths are uncorrelated to each other, the
ranged facet provides no constraint on the features outside the facet. As soon as the platform starts moving,
visual measurements begin to correlate all feature depths, and the depths of all features become observable
from a single ranged facet.

\section{Experimental Setup} 
\label{sec:testing}

We recorded two datasets to characterize the performance of our approach. The first one is an outdoor flight
in an urban environment, which is a common scenario in aerial robotics. The second sequence is aimed at stressing the facet
assumption in an indoor environment, where full-state ground truth based on motion capture is available. Both tests consist of a uniform-velocity
traverse, i.e. a straight line at constant speed, since this is the most limiting unobservable direction of VIO.
Processing was done offline for this paper since the focus is on analysis, but we previously demonstrated the
real-time performance of the range-visual-inertial odometry algorithm at 30 frames per seconds on a Snapdragon 820 processor~\cite{delaune2020}, which
is the space hardware baseline for the next Mars helicopter mission concept.

\subsection{Sensors}

Range data was provided by a Garmin Lidar Lite V3 single-point static laser range finder. This can range
up to 40 m with a 2.5-cm accuracy, weighs only 22 g, and is less than 5 cm-long. The monocular navigation
camera was a global-shutter Omnivision OV7251, providing $640\times480$ \mbox{8-bit} grayscale images in auto-exposure
mode. Inertial data was delivered by a STIM30
Code
Related Papers
About arXivLabs
0 tactical-grade IMU. The camera was collecting data
at 30 Hz, the LRF at 25 Hz, and the IMU at 250 Hz. The sensor suite was mounted
on a rigid platform. The camera intrinsics and extrinsincs were calibrated, including the angles between the camera
and LRF optical axes. The distance between the camera and the LRF was neglected, as the two sensors were mounted side
by side.

\subsection{Outdoor Test}

\subsubsection{Flight Sequence}

For this test, the sensor platform was mounted on a GPS-controlled hexacopter. After take-off, the rotorcraft ascended
to a cruise altitude of 11 m, and initiated a 150~m straight horizontal traverse at a constant speed of 2 m/s. The traverse was controlled within the performance limits of the on-board Pixhawk 2.1 Cube autopilot set up with the ArduCopter APM firmware. No
inertial excitation was provided before take-off.

The flight path was chosen so that the first half of the trajectory covers flat ground, where the facet assumption is
likely to be respected; and the second half is over buildings with non-flat roofs, including structure discontinuities where
the facet assumption may be challenged. Our video supplement includes the full sequence, and a height profile to illustrate
structure variations along the flight path. Figure~\ref{fig:delaunay} shows the image
facets at the transition between flat ground and rooftops. 

\subsubsection{GPS Ground Truth}

Position ground truth was provided by a RTK-GPS system composed of a Trimble BD930-UHF receiver with a BX982 base station.
This system provides centimeter accuracy in clear outdoor environments. It also served as time server for our logging computer,
so sensor data was readily time-synchronized with ground truth.

We attempted to run a GPS-IMU filter to get attitude ground truth, in addition to position only. However the horizontal
accelerations were too small to observe the heading component of attitude~\cite{weiss2012phd}, which ended up drifting.

\subsection{Indoor Test}
\label{sub:indoor-seq}

\subsubsection{Hand-Held Sequence}

Our second data set was recorded in an indoor environment, with the intention to create a stress test for the facet
assumption. We arranged boxes of different heights next to each other in straight line under the LRF path, to create a structure
with multiple 90$^\circ$ drop-offs, which result in severe relative height changes when the sensors are in close proximity to
the top of the boxes. We refer the reader our video material to observe the violations of the facet assumption, which happens
frequently in this sequence.

To highlight the benefit of the ranged facet model, we also made it a stress case for VIO: the sensor
platform was hand-held to smooth the motion and eliminate the residual accelerations coming from the hexacopter controls\footnote{Residual
accelerations from hand motion are still present though.}; the environment had surfaces with little texture to limit long feature
tracks and increase visual scale drift; and once again, no inertial excitation was provided before the horizontal traverse.

\subsubsection{Motion Capture Ground Truth}

Another motivation for an indoor dataset was to obtain complete and accurate ground truth to fully characterize
range-VIO performance in a stress case. Our test arena was equipped with 10 Vicon Vero motion capture cameras, which
typically provide millimeter accuracy in position and sub-degree in orientation. For velocity ground truth, we filtered
Vicon pose measurement with the on-board IMU~\cite{weiss2011icra}. We should note that this ground truth is not fully independent
from the state estimates since the same IMU was used for both.
\section{Results} 
\label{sec:results}
This section discusses the performance of our range-VIO algorithm on the sequences presented in the previous section. The
visual state was set to accommodate $M=4$ poses in the sliding window, and $N=27$ SLAM features. The Mahalanobis distance
test to capture outliers was set to $2\sigma$, with $\sigma$ the estimated range standard deviation. VIO was run with the exact
same settings as range-VIO in all our comparison tests. The only difference was the additional processing of the range measurement
with the ranged facet model in range-VIO.

\subsection{Outdoor Flight Tests}

%

\begin{figure}[!b]
\centering
\vspace{-14pt}
\begin{subfigure}{.35\textwidth}
  \centering
  \includegraphics[width=\linewidth]{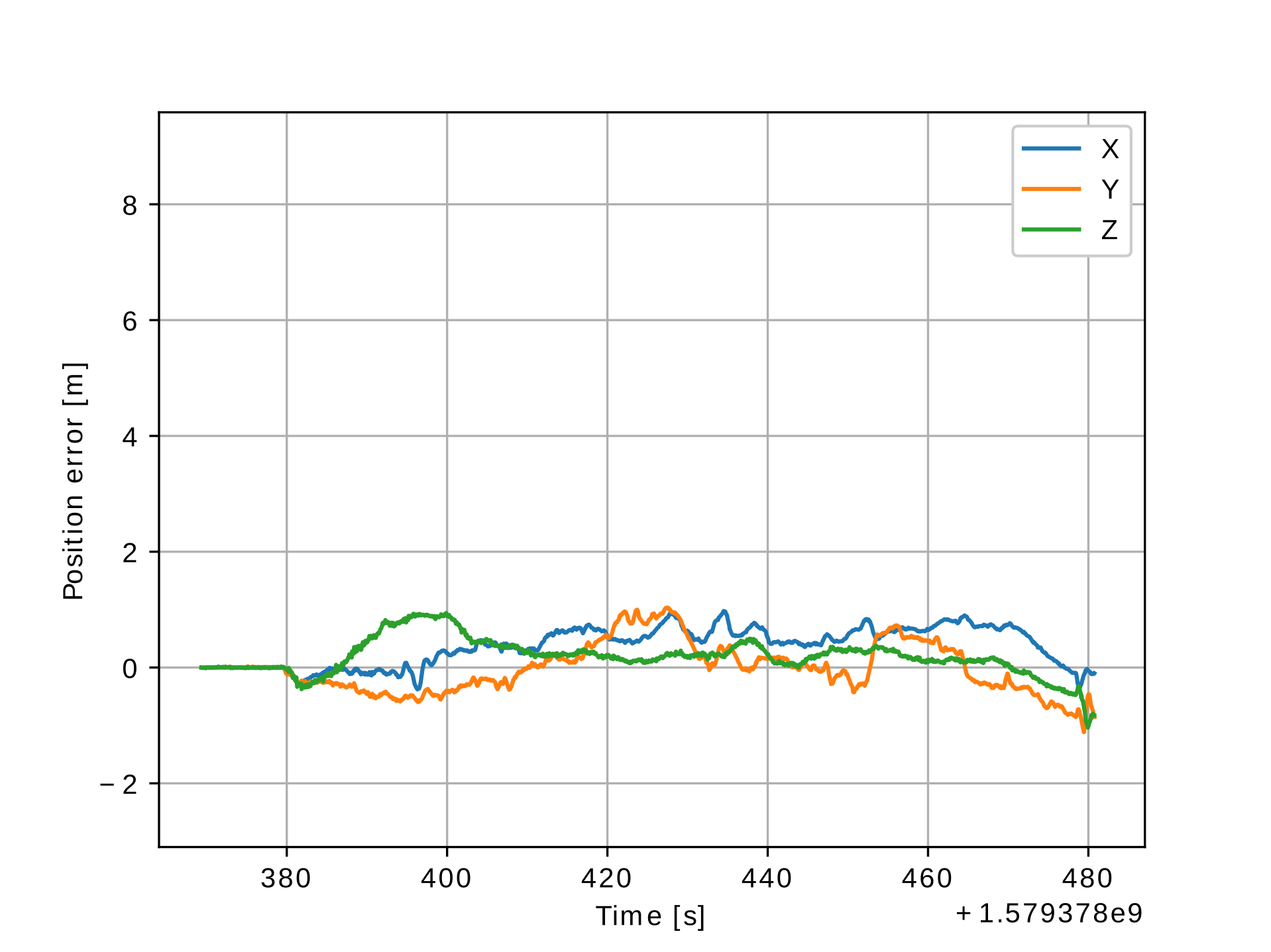}  
  \caption{Range-VIO position error}
  \label{fig:poserr-outdoor-rvio}
\end{subfigure}
\begin{subfigure}{.35\textwidth}
  \centering
  \includegraphics[width=\linewidth]{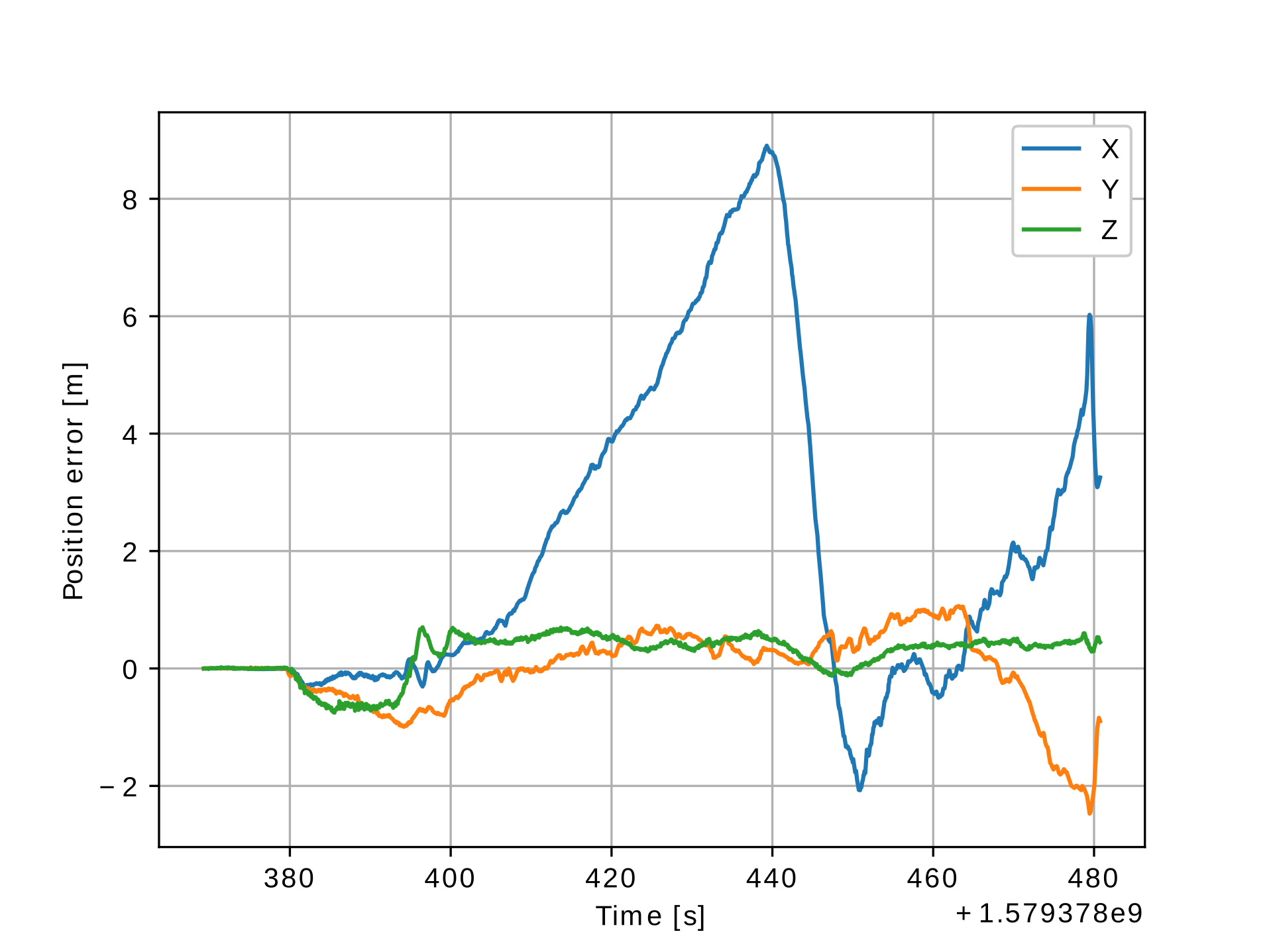}  
  \caption{VIO position error}
  \label{fig:poserr-outdoor-vio}
\end{subfigure}
\caption{Position errors for range-VIO (top) and
VIO (bottom) on the outdoor dataset. The $X$ and $Y$ axes are horizontal,
$Z$ is up. $X$ was aligned with the direction of the traverse.}
\label{fig:poserr-outdoor}
\end{figure}

Figure~\ref{fig:poserr-outdoor} compares the position errors of range-VIO and VIO during the outdoor traverse.
Range-VIO maximum errors remain below 1 m on each axis, which is under $0.6\%$ of the distance travelled. This performance
is similar to state-of-the-art VIO under excitation~\cite{delmerico2018icra}. Conversely, the VIO
error rises along the direction of the traverse ($X$ axis) from the time it is initiated, and up to values
9 times larger compared to range-VIO.
\begin{figure*}[t]
\begin{subfigure}{.33\textwidth}
  \centering
  \includegraphics[width=\linewidth]{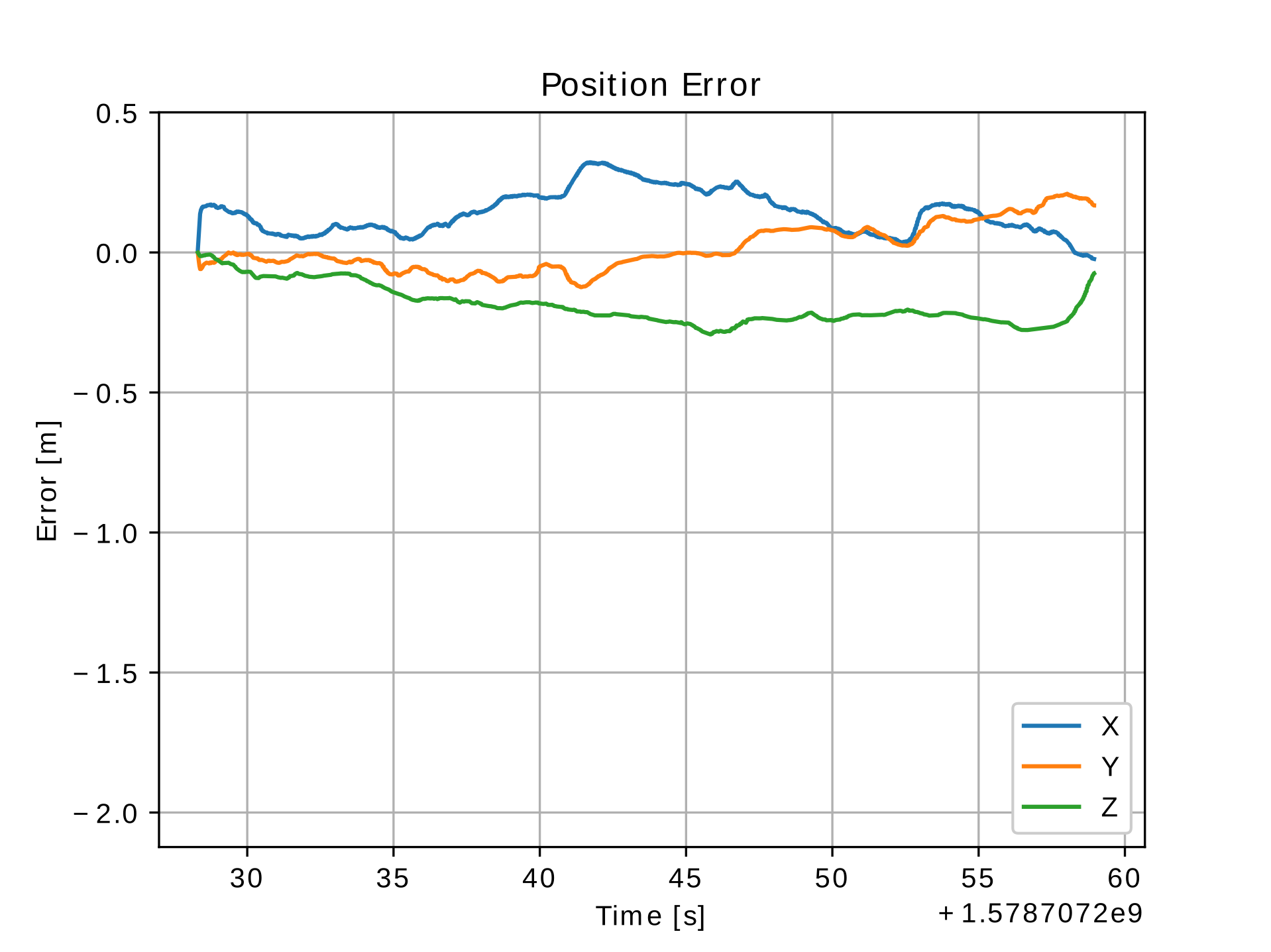}  
  \caption{Range-VIO position error}
\end{subfigure}
\begin{subfigure}{.33\textwidth}
  \centering
  \includegraphics[width=\linewidth]{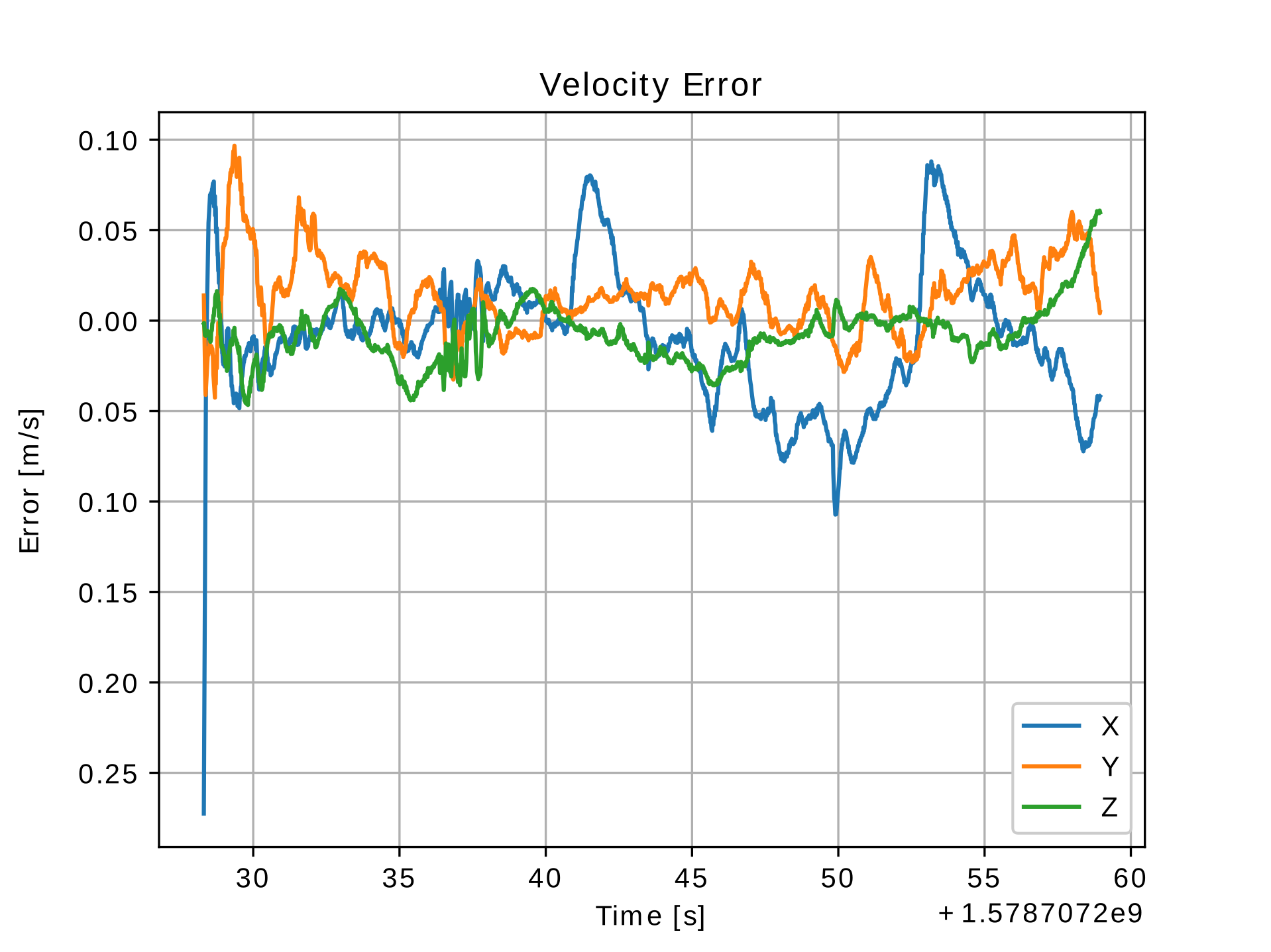}  
  \caption{Range-VIO velocity error}
\end{subfigure}
\begin{subfigure}{.33\textwidth}
  \centering
Code
Related Papers
About arXivLabs

  \includegraphics[width=\linewidth]{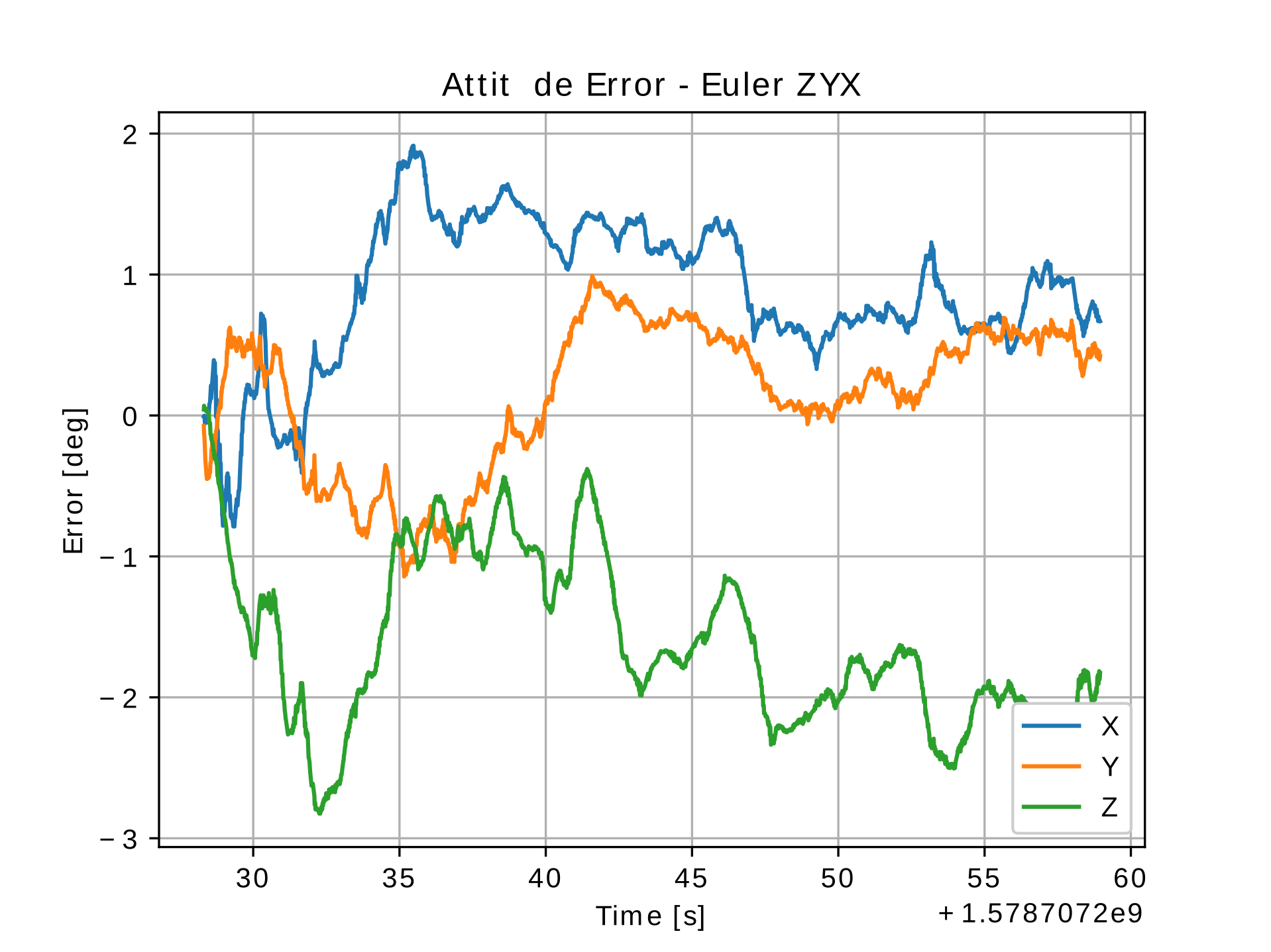}  
  \caption{Range-VIO attitude error}
\end{subfigure}
\begin{subfigure}{.33\textwidth}
  \centering
  \includegraphics[width=\linewidth]{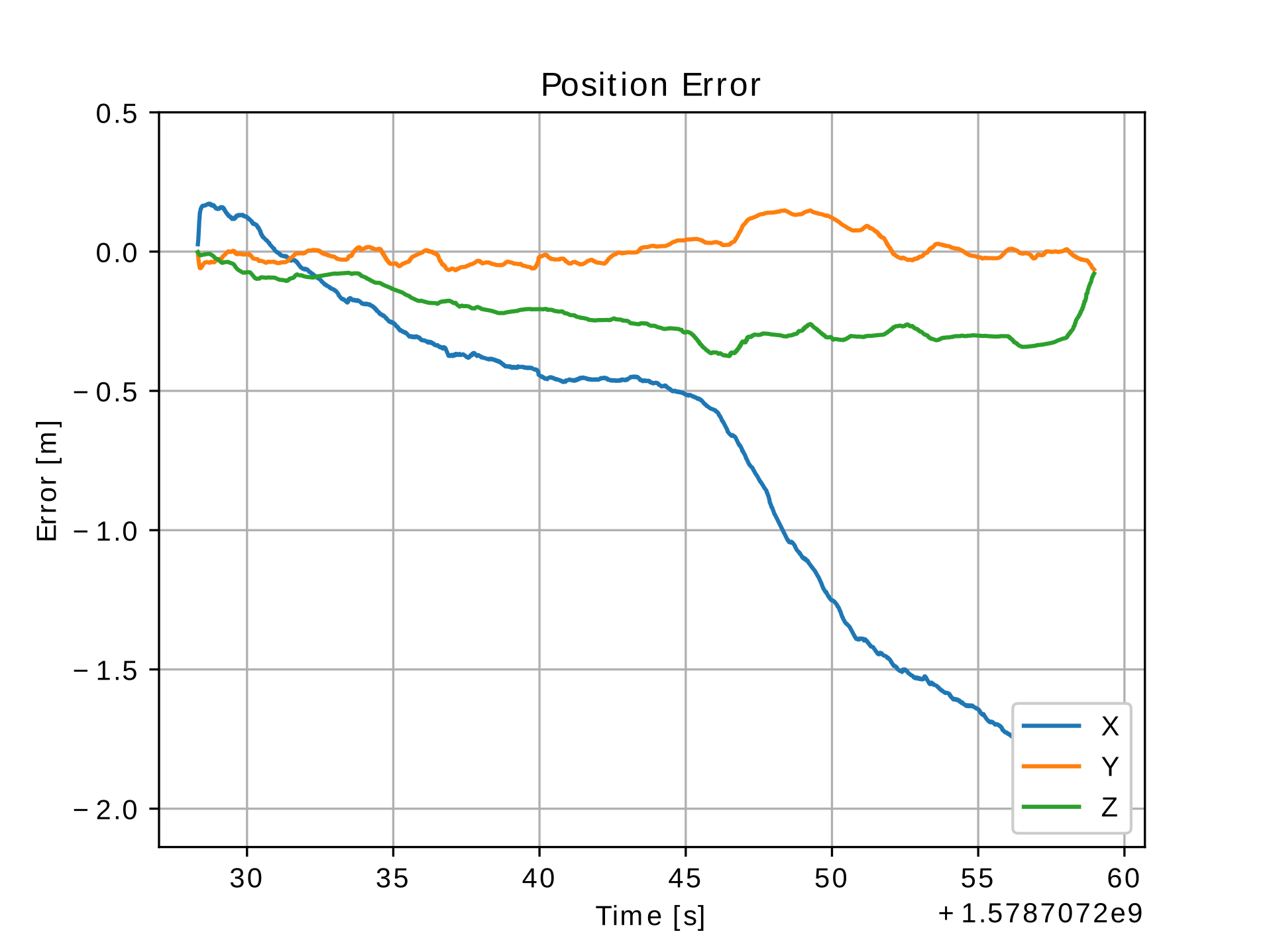}  
  \caption{VIO position error}
\end{subfigure}
\begin{subfigure}{.33\textwidth}
  \centering
  \includegraphics[width=\linewidth]{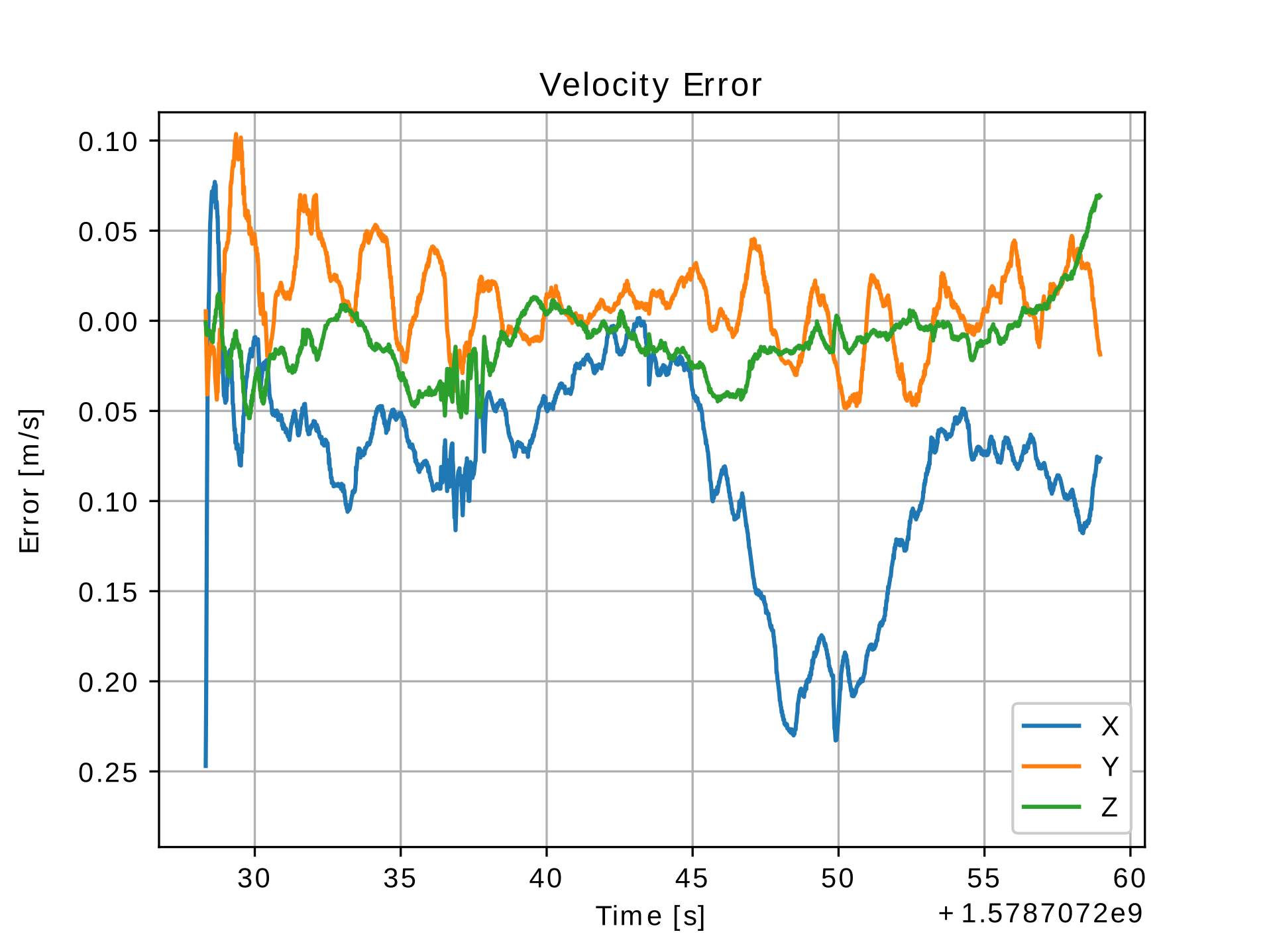}  
  \caption{VIO velocity error}
\end{subfigure}
\begin{subfigure}{.33\textwidth}
  \centering
  \includegraphics[width=\linewidth]{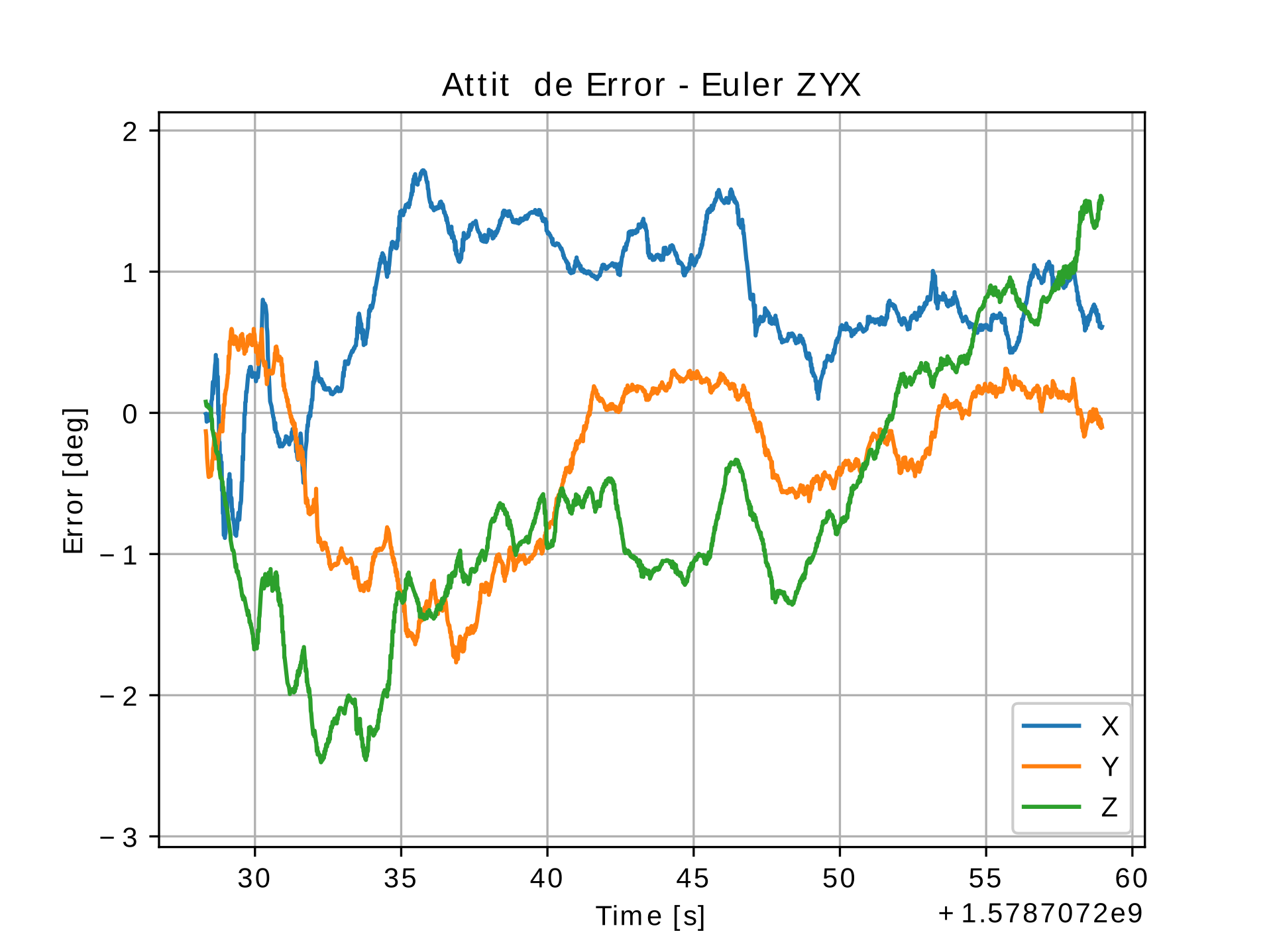}  
  \caption{VIO attitude error}
\end{subfigure}
\caption{Position (left), velocity (center) and attitude (right) errors for
range-VIO (top) and VIO (bottom) on the indoor stress dataset. The $X$ and $Y$ axes are horizontal, $Z$ is up.
$X$ was aligned with the direction of the traverse.}
\label{fig:state-err-indoor}
\end{figure*}

We note that the VIO error is consistent with a scale error, which is not observable for VIO under the
constant-acceleration traverse. This is a clear illustration of the observability benefit
of range-VIO on a trajectory commonly used in robotics. We also note that the range-VIO errors in
Figure~\ref{fig:poserr-outdoor}(a) do not suffer from the transition between a flat terrain and a 3D
structure, that occurs at $t=425$~s. This is a good indication
that the facets constructed with real-world visual features efficiently capture the structure of the scene.
Additionally a 7-m ranged facet outlier occurred at $t=410$~s, as
the LRF hits a street light. This can be observed in the range profile shown in our video material. However it does not
affect the range-VIO estimates in Figure~\ref{fig:poserr-outdoor}(a),
showing the efficiency of our range outlier rejection scheme.

\subsection{Indoor Stress Tests}

To further assess the robustness of the facet model, the indoor sequence discussed in
Subsection~\ref{sub:indoor-seq} was used as a stress case. Figure~\ref{fig:state-err-indoor}
compares range-VIO and VIO errors in position, velocity and orientation, since ground truth
was available for all these states indoors.
The scale drift reduction is clearly visible along the direction of travel in
Figures~\ref{fig:state-err-indoor}(a) and \ref{fig:state-err-indoor}(d). Range-VIO has a maximum
position error of 30~cm, or $2.5\%$ while VIO errors grow to 2 m, or $17\%$ in these challenging
visual conditions and without excitation.

The velocity and orientation plots are a good illustration of how the facet assumption can work over challenging environments.
Figure~\ref{fig:state-err-indoor}(b) and \ref{fig:state-err-indoor}(e) show the velocity errors benefit from scale observability
in range-VIO, since they are up to twice lower than VIO. Likewise, range-VIO orientation errors in Figure~\ref{fig:state-err-indoor}(c)
slightly differ from that of VIO in Figure~\ref{fig:state-err-indoor}(f), especially in the Z (yaw) axis. We interpret these differences
as error accumulation cause by ranged facet assumption violations too small to be caught by the Mahalanobis
range outlier rejection. This only happens around the global yaw axis, which is unobservable to both methods. However, even in this extreme stress
case, the yaw error has the same order of magnitude between range-VIO and VIO, while range-VIO clearly outperforms VIO
in terms of position and velocity drift reduction.

Finally, we refer the reader to our video material for additional comparison
results in a sequence with large excitation and good visual texture. Under these optimal conditions, VIO performs on par
with range-VIO, having a maximum position error of 40 cm. This confirms that VIO was not detuned previously, but only suffered
from the lack of excitation. It also confirms that range-VIO does not degrade VIO performance in the presence of good excitation and visual conditions.

\section{Conclusion} 
\label{sec:conclusion}

VIO-based robotic applications are limited by the inability to observe scale without excitation. In aerial robotics, scale observability without excitation is critical for even the
most basic hovering and straight-line trajectories. Our main interest is for aerial exploration of distant worlds,
like Mars~\cite{bayard2019scitech}. Common terrestrial applications include conditions where GPS is unavailable (defense, underground),
degraded (tall buildings, canyons), or not accurate enough (indoors). 

 Using a simple 1D laser range finder, our range-VIO approach eliminates scale drift in the absence of excitation while retaining the minimal size, weight and power requirements of VIO.
 A theoretical analysis demonstrated the observability of scale in such conditions. Results on constant-velocity real flight data showed error reduction by a factor of 9 compared to VIO.
 
 The novel range update is based on a facet scene assumption that efficiently leverages VIO feature depth estimates to handle unknown structures. The facets can scale from a flat world assumption, to virtually no structure assumption at all based on visual feature density. This paper and supplement report~\cite{delaune2020b} provide a full derivation of the range-VIO model. Range-VIO does not require additional states with respect to VIO, and does not add significant computational cost. We demonstrated the robustness of our facet assumption in a stress case.

Future extensions include increasing visual feature density around the LRF impact point on the scene to improve accuracy further. We also investigate the use of magnetometers and sun sensors to address the next major unobservable direction:
orientation about the gravity vector.


\bibliographystyle{unsrt}
\bibliography{root}

\begin{thebibliography}{10}

\bibitem{bayard2019scitech}
D.~S. Bayard, D.~T. Conway, R.~Brockers, J.~Delaune, L.~Matthies, H.~Grip,
  G.~Merewether, T.~Brown, and A.~San~Martin.
\newblock
  \href{https://arc.aiaa.org/doi/pdfplus/10.2514/6.2019-1411}{Vision-Based
  Navigation for the NASA Mars Helicopter}.
\newblock In {\em AIAA Scitech Forum}, 2019.

\bibitem{delaune2020}
J.~Delaune, R.~Brockers, D.~S. Bayard, H.~Dor, R.~Hewitt, J.~Sawoniewicz,
  G.~Kubiak, T.~Tzanetos, L.~Matthies, and J.~Balaram.
\newblock \href{https://ieeexplore.ieee.org/document/9172289}{Extended
  Navigation Capabilities for a Future Mars Science Helicopter Concept}.
\newblock In {\em IEEE Aerospace Conference}, 2020.

\bibitem{delaune2020b}
J.~Delaune, D.S. Bayard, and R.~Brockers.
\newblock \href{https://arxiv.org/abs/2010.06677}{xVIO: A Range-Visual-Inertial
  Odometry Framework}.
\newblock Technical report, Jet Propulsion Laboratory, 2020, arXiv:2010.06677.

\bibitem{weiss2012icra}
S.~Weiss, M.~W. Achtelik, S.~Lynen, M.~Chli, and R.~Siegwart.
\newblock
  \href{https://ieeexplore.ieee.org/abstract/document/6225147}{Real-time
  Onboard Visual-Inertial State Estimation and Self-Calibration of MAVs in
  Unknown Environments}.
\newblock In {\em International Conference on Robotics and Automation (ICRA)},
  pages 957--964. IEEE, 2012.

\bibitem{klein2007ismar}
G.~Klein and D.~Murray.
\newblock
  \href{http://www.robots.ox.ac.uk/~gk/publications/KleinMurray2007ISMAR.pdf}{Parallel
  Tracking and Mapping for Small {AR} Workspaces}.
\newblock In {\em Proc. Sixth {IEEE} and {ACM} International Symposium on Mixed
  and Augmented Reality {(ISMAR'07)}}, 2007.

\bibitem{forster2014icra}
C.~Forster, M.~Pizzoli, and D.~Scaramuzza.
\newblock \href{http://rpg.ifi.uzh.ch/docs/ICRA14_Forster.pdf}{{SVO}: Fast
  Semi-Direct Monocular Visual Odometry}.
\newblock In {\em International Conference on Robotics and Automation (ICRA)}.
  IEEE, 2014.

\bibitem{murTRO2015}
R.~Mur-Artal, J.~M.~M. Montiel, and Juan~D. Tard\'os.
\newblock \href{https://arxiv.org/abs/1502.00956}{{ORB-SLAM}: a Versatile and
  Accurate Monocular {SLAM} System}.
\newblock {\em IEEE Transactions on Robotics}, 31(5):1147--1163, 2015.

\bibitem{engel2018pami}
J.~Engel, V.~Koltun, and D.~Cremers.
\newblock
  \href{https://vision.in.tum.de/_media/spezial/bib/engel_et_al_pami2018.pdf}{Direct
  Sparse Odometry}.
\newblock {\em IEEE Transactions on Pattern Analysis and Machine Intelligence},
  March 2018.

\bibitem{leutenegger2014ijrr}
S.~Leutenegger, S.~Lynen, M.~Bosse, R.~Siegwart, and P.~Furgale.
\newblock
  \href{https://spiral.imperial.ac.uk:8443/bitstream/10044/1/23413/2/ijrr2014_revision_1.pdf}{Keyframe-Based
  Visual-Inertial Odometry Using Nonlinear Optimization}.
\newblock {\em The International Journal of Robotics Research}, 34, 02 2014.

\bibitem{mourikis2007icra}
A.~I. Mourikis and S.~I. Roumeliotis.
\newblock \href{https://ieeexplore.ieee.org/document/4209642}{A multi-state
  constraint Kalman filter for vision-aided inertial navigation}.
\newblock In {\em IEEE International Conference on Robotics and Automation
  (ICRA)}, 2007.

\bibitem{Bloesch2015}
M.~Bloesch, S.~Omari, M.~Hutter, and R.~Siegwart.
\newblock \href{https://ieeexplore.ieee.org/document/7353389}{Robust Visual
  Inertial Odometry Using a Direct EKF-Based Approach}.
\newblock In {\em IEEE/RSJ International Conference on Intelligent Robots and
  Systems (IROS)}, 2015.

\bibitem{forster2017ieee}
C.~Forster, L.~Carlone, F.~Dellaert, and D.~Scaramuzza.
\newblock \href{http://rpg.ifi.uzh.ch/docs/TRO16_forster.pdf}{On-Manifold
  Preintegration for Real-Time Visual-Inertial Odometry}.
\newblock {\em IEEE Transactions on Robotics}, 33(1):1–21, 2017.

\bibitem{Qin2017}
T.~Qin, P.~Li, and S.~Shen.
\newblock \href{https://arxiv.org/pdf/1708.03852.pdf}{VINS-Mono: A Robust and
  Versatile Monocular Visual-Inertial State Estimator}.
\newblock {\em IEEE Transactions on Robotics}, 34(4):1004--1020, 2018.

\bibitem{Stumberg2018}
L.~von Stumberg, V.~C. Usenko, and D.~Cremers.
\newblock \href{https://arxiv.org/abs/1804.05625}{Direct Sparse Visual-Inertial
  Odometry Using Dynamic Marginalization}.
\newblock {\em IEEE International Conference on Robotics and Automation
  (ICRA)}, pages 2510--2517, 2018.

\bibitem{delmerico2018icra}
J.~A. Delmerico and D.~Scaramuzza.
\newblock \href{http://rpg.ifi.uzh.ch/docs/ICRA18_Delmerico.pdf}{A Benchmark
  Comparison of Monocular Visual-Inertial Odometry Algorithms for Flying
  Robots}.
\newblock {\em International Conference on Robotics and Automation (ICRA)},
  pages 2502--2509, 2018.

\bibitem{martinelli2011tr}
A.~Martinelli.
\newblock \href{https://hal.inria.fr/inria-00569083v1/document}{Closed-Form
  Solutions for Attitude, Speed, Absolute Scale and Bias Determination by
  Fusing Vision and Inertial Measurements}.
\newblock Technical report, INRIA, 2011.

\bibitem{eagle2011ijrr}
E.~S. Jones and S.~Soatto.
\newblock
  \href{https://journals.sagepub.com/doi/10.1177/0278364910388963#articleCitationDownloadContainer}{Visual-inertial
  navigation, mapping and localization: A scalable real-time causal approach}.
\newblock {\em The International Journal of Robotics Research}, 30(4):407--430,
  2011.

\bibitem{kelly2011ijrr}
J.~Kelly and G.~S. Sukhatme.
\newblock
  \href{https://journals.sagepub.com/doi/10.1177/0278364910388963}{Visual-Inertial
  Sensor Fusion: Localization, Mapping and Sensor-to-Sensor Self-calibration}.
\newblock {\em The International Journal of Robotics Research}, 30(1):56--79,
  2011.

\bibitem{li2013ijrr}
M.~Li and A.~I. Mourikis.
\newblock
  \href{https://journals.sagepub.com/doi/10.1177/0278364913481251}{High-precision,
  consistent EKF-based visual-inertial odometry}.
\newblock {\em The International Journal of Robotics Research}, 32(6):690--711,
  2013.

\bibitem{heschtro2014}
J.~A. Hesch, D.~G. Kottas, S.~L. Bowman, and S.~I. Roumeliotis.
\newblock \href{https://ieeexplore.ieee.org/document/6605544}{Consistency
  Analysis and Improvement of Vision-aided Inertial Navigation}.
\newblock {\em IEEE Transactions on Robotics}, 30(1):158--176, 2014.

\bibitem{wu2016minn}
K.~J. Wu and S.~I. Roumeliotis.
\newblock
  \href{https://pdfs.semanticscholar.org/4947/d28474cddd5cba7f592a8d3dcfd63316c02e.pdf?_ga=2.146440563.630791678.1580088558-163640800.1579748327}{Unobservable
  Directions of VINS Under Special Motions}.
\newblock Technical report, University of Minnesota, 2016.

\bibitem{kottas2013iros}
D.~G. Kottas, K.~J. Wu, and S.~I. Roumeliotis.
\newblock
  \href{https://www-users.cs.umn.edu/~stergios/papers/IROS_2013_Hover_Dimitris_Kejian.pdf}{Detecting
  and dealing with hovering maneuvers in vision-aided inertial navigation
  systems}.
\newblock In {\em 2013 IEEE/RSJ International Conference on Intelligent Robots
  and Systems}, pages 3172--3179, 2013.

\bibitem{mohamed2019ieee}
S.~Mohamed, M.~Haghbayan, T.~Westerlund, J.~Heikkonen, H.~Tenhunen, and
  J.~Plosila.
\newblock
  \href{https://ieeexplore.ieee.org/abstract/document/8764393/authors#authors}{A
  Survey on Odometry for Autonomous Navigation Systems}.
\newblock {\em IEEE Access}, PP:1--1, 07 2019.

\bibitem{angladon2017jrtip}
V.~Angladon, S.~Gasparini, V.~Charvillat, T.~Pribanic, T.~Petković, M.~Donlic,
  B.~Ahsan, and F.~Bruel.
\newblock \href{https://hal.archives-ouvertes.fr/hal-01654706/document}{An
  evaluation of real-time RGB-D visual odometry algorithms on mobile devices}.
\newblock {\em Journal of Real-Time Image Processing}, 02 2017.

\bibitem{taketomi2017cva}
T.~Taketomi, H.~Uchiyama, and S.~Ikeda.
\newblock
  \href{https://link.springer.com/article/10.1186/s41074-017-0027-2}{Visual
  SLAM algorithms: a survey from 2010 to 2016}.
\newblock {\em IPSJ Transactions on Computer Vision and Applications}, 9:1--11,
  2017.

\bibitem{urzua2017mav}
S.~Urzua, R.~Munguía, and A.~Grau.
\newblock
  \href{https://journals.sagepub.com/doi/full/10.1177/1756829317705325}{Vision-based
  SLAM system for MAVs in GPS-denied environments}.
\newblock {\em International Journal of Micro Air Vehicles}, 9(4):283--296,
  2017.

\bibitem{weiss2011icra}
S.~Weiss and R.~Siegwart.
\newblock \href{https://ieeexplore.ieee.org/document/5979982}{Real-Time Metric
  State Estimation for Modular Vision-Inertial Systems}.
\newblock In {\em International Conference on Robotics and Automation (ICRA)}.
  IEEE, 2012.

\bibitem{trawny2005minn}
N.~Trawny and S.~I. Roumeliotis.
\newblock \href{http://mars.cs.umn.edu/tr/reports/Trawny05b.pdf}{Indirect
  Kalman Filter for 3D Attitude Estimation}.
\newblock Technical report, University of Minnesota, 2005.

\bibitem{li2012rss}
M.~Li and A.~Mourikis.
\newblock
  \href{http://www.roboticsproceedings.org/rss08/p31.pdf}{Optimization-Based
  Estimator Design for Vision-Aided Inertial Navigation}.
\newblock In {\em Robotics: Science and Systems conference}, 2012.

\bibitem{civera2008tro}
J.~Civera, A.~Davison, and J.~Montiel.
\newblock
  \href{https://www.doc.ic.ac.uk/~ajd/Publications/civera_etal_tro2008.pdf}{Inverse
  Depth Parametrization for Monocular SLAM}.
\newblock {\em IEEE Transactions on Robotics}, 24(5):932--945, 2008.

\bibitem{delaune2019iros}
J.~Delaune, R.~Hewitt, L.~Lytle, C.~Sorice, R.~Thakker, and L.~Matthies.
\newblock \href{https://ieeexplore.ieee.org/document/8968238}{Thermal-Inertial
  Odometry for Autonomous Flight Throughout the Night}.
\newblock In {\em International Conference on Intelligent Robots and Systems
  (IROS)}. IEEE/RSJ, 2019.

\bibitem{rosten2010faster}
E.~Rosten, R.~Porter, and T.~Drummond.
\newblock \href{https://arxiv.org/pdf/0810.2434.pdf}{Faster and better: A
  machine learning approach to corner detection}.
\newblock {\em Transactions on Pattern Analysis and Machine Intelligence},
  32(1):105--119, 2010.

\bibitem{Bouguet2000}
J.-Y. Bouguet.
\newblock
  \href{http://robots.stanford.edu/cs223b04/algo_tracking.pdf}{Pyramidal
  implementation of the Lucas Kanade feature tracker}.
\newblock {\em Intel Corporation, Microprocessor Research Labs}, 2000.

\bibitem{shi1994}
J.~Shi and C.~Tomasi.
\newblock \href{https://ieeexplore.ieee.org/document/323794}{Good Features to
  Track}.
\newblock In {\em Conference on Computer Vision and Pattern Recognition
  (CVPR)}, pages 594--600. IEEE, 1994.

\bibitem{fischler1981acm}
M.~A. Fischler and R.~C. Bolles.
\newblock \href{https://dl.acm.org/doi/10.1145/358669.358692}{Random sample
  consensus: a paradigm for model fitting with applications to image analysis
  and automated cartography}.
\newblock {\em Communications of the ACM}, 24(6):381--395, 1981.

\bibitem{gartner2013ETH}
B.~G\"{a}rtner and M.~Hoffmann.
\newblock
  \href{https://www.ti.inf.ethz.ch/ew/Lehre/CG13/lecture/cg-2013.pdf}{Computational
  Geometry Lecture Notes HS 2013}.
\newblock Technical report, ETH Z\"{u}rich, 2013.

\bibitem{hesch2012minn}
J.~A. Hesch, D.~G. Kottas, S.~L. Bowman, and S.~I. Roumeliotis.
\newblock
  \href{https://www.seas.upenn.edu/~seanbow/papers/tr_2012_1.pdf}{Observability-constrained
  Vision-aided Inertial Navigation}.
\newblock Technical report, University of Minnesota, 2012.

\bibitem{weiss2012phd}
S.~Weiss.
\newblock {\em
  \href{http://e-collection.library.ethz.ch/eserv/eth:5889/eth-5889-02.pdf}{Vision
  Based Navigation for Micro Helicopters}}.
\newblock PhD thesis, ETH Z\"{u}rich, 2012.

\end{thebibliography}

\end{document}